\documentclass[10pt,twocolumn,letterpaper]{article}

\usepackage{iccv}
\usepackage{times}
\usepackage{epsfig}
\usepackage{graphicx}
\usepackage{amsmath}
\usepackage{amssymb}
\usepackage{wrapfig}
\usepackage{tabularx}
\usepackage{enumitem}

\usepackage{caption}
\DeclareMathOperator*{\rint}{\ThisStyle{\rotatebox{10}{$\SavedStyle\!\int\!$}}}
\usepackage{scalerel}

\usepackage{booktabs}
\newcommand{\ra}[1]{\renewcommand{\arraystretch}{#1}}
\newcolumntype{F}{>{\centering\arraybackslash}p{1em}}
\newcolumntype{C}{>{\centering\arraybackslash}p{6em}}

\newcommand{\nickname}{GRF}

\usepackage[pagebackref=true,breaklinks=true,letterpaper=true,colorlinks,bookmarks=false]{hyperref}

\iccvfinalcopy 

\def\iccvPaperID{6616} 

\ificcvfinal\fi

\begin{document}

\title{GRF: Learning a General Radiance Field for 3D Representation and Rendering }

\author{Alex Trevithick\\
UC San Diego, University of Oxford\\
{\tt\small atrevithick@ucsd.edu}
\and
Bo Yang\\
vLAR Group, The Hong Kong Polytechnic University\\
{\tt\small bo.yang@polyu.edu.hk}
}

\maketitle
\ificcvfinal\thispagestyle{empty}\fi

\begin{abstract}
We present a simple yet powerful neural network that implicitly represents and renders 3D objects and scenes only from 2D observations. The network models 3D geometries as a general radiance field, which takes a set of 2D images with camera poses and intrinsics as input, constructs an internal representation for each point of the 3D space, and then renders the corresponding appearance and geometry of that point viewed from an arbitrary position. The key to our approach is to learn local features for each pixel in 2D images and to then project these features to 3D points, thus yielding general and rich point representations. We additionally integrate an attention mechanism to aggregate pixel features from multiple 2D views, such that visual occlusions are implicitly taken into account. Extensive experiments demonstrate that our method can generate high-quality and realistic novel views for novel objects, unseen categories and challenging real-world scenes.
\end{abstract}

\section{Introduction}
\label{sec:intro}
Understanding the precise 3D structure of a real-world environment and realistically re-rendering it from free viewpoints is a key enabler for many critical tasks, ranging from robotic manipulation to augmented reality. Classic approaches to recover the 3D geometry mainly include the structure from motion (SfM) \cite{Ozyesil2017} and simultaneous localization and mapping (SLAM) \cite{Cadena2016} pipelines. However, they can only reconstruct sparse and discrete 3D point clouds which are unable to contain geometric details. 

The recent advances in deep neural networks have yielded rapid progress in 3D modeling. Most of them focus on the explicit 3D shape representations such as voxel grids \cite{Chan2016}, point clouds \cite{Fan2017}, and triangle meshes \cite{Wang2018f}. However, these representations are discrete and sparse, limiting the recovered 3D structures to extremely low spatial resolution. In addition, these networks usually require large-scale 3D shapes for supervision, resulting in the trained models over-fitting particular datasets and lacking generalization to novel geometries.

\begin{figure}[t]
\setlength{\abovecaptionskip}{ 4 pt}
\setlength{\belowcaptionskip}{ -12 pt}
\centering
   \includegraphics[width=1\linewidth]{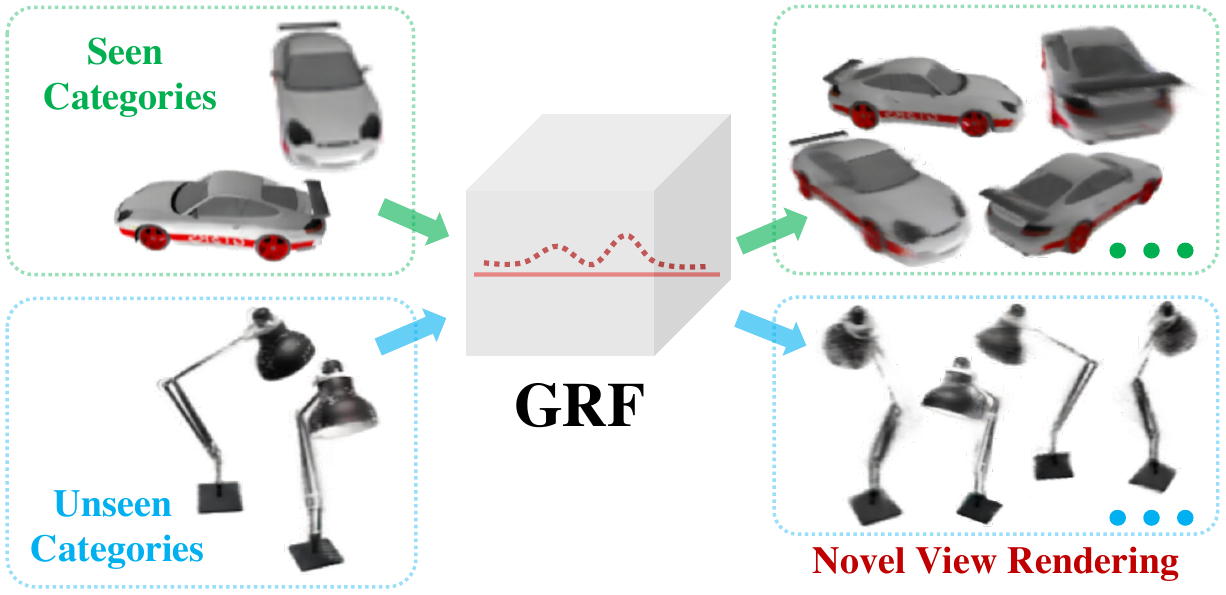}
\caption{A single model of our \nickname{} infers high-quality novel views for new objects of seen and unseen categories, demonstrating its strong capability for 3D representation and rendering.}
\label{fig:grf_intro}
\end{figure}

Encoding geometries into multilayer perceptrons(MLPs) \cite{Mescheder2019,Park2019} recently emerges as a promising direction in 3D reconstruction from 2D images. Its key advantage is the ability to model 3D structures continuously instead of discretely, and therefore it has the potential to achieve unlimited spatial resolution in theory. However, many of these methods require 3D geometry for supervision to learn the 3D shapes from images. By introducing a recurrent neural network based renderer, SRNs \cite{Sitzmann2019} is among the early work to learn implicit surface representations only from 2D images, but it renders over-smoothed images without details. Alternatively, by leveraging the volume rendering to synthesize new views with 2D supervision, the very recent NeRF \cite{Mildenhall2020} directly encodes the 3D structure into a radiance field via MLPs, achieving an unprecedented level of fidelity.

Nevertheless, NeRF has two major limitations: 1) since 3D content is encoded into the weights of an MLP, the trained network (\textit{i.e.,} a learned radiance field) can only represent a single structure, and is unable to generalize across novel geometries; and 2) because the shape and appearance of each spatial 3D location along a light ray is only optimized by individual pixel RGBs, the learned representations of that location do not have rich geometric patterns, resulting in less photo-realistic rendered images.

In this paper, we propose a \textbf{g}eneral \textbf{r}adiance \textbf{f}ield (\nickname{}), a simple yet powerful neural network that builds upon NeRF \cite{Mildenhall2020}, overcoming these two limitations. Our \nickname{} takes a set of 2D images with camera poses and intrinsics, a 3D query point, and its query viewpoint (\textit{i.e.}, the camera location $xyz$) as input, and predicts the RGB value and volumetric density at that query point. Our network learns to represent  3D content from sparse 2D observations, and to infer shape and appearance from previously unobserved viewing angles. Note that the inferred shape and appearance of any particular 3D query point explicitly takes into account its local geometry from the available 2D observations.
In particular, our proposed \nickname{} consists of four components:
\begin{itemize}[leftmargin=*,topsep=0pt]
\itemsep-0.2em
    \item Extracting general 2D visual features for every light ray from the input 2D observations;
    \item Reprojecting the corresponding 2D features back to the query 3D point using multi-view geometry;
    \item Aggregating all reprojected features of the query point with attention, where visual occlusions are implicitly considered;
    \item Rendering the aggregated features of the query 3D point along a particular query viewpoint, and producing the corresponding RGB and volumetric density via NeRF \cite{Mildenhall2020}.
\end{itemize}

These four components enable our \nickname{} to distinguish itself from existing approaches: 1) Compared with the classic SfM/SLAM systems, our \nickname{} can represent the 3D content with continuous surfaces; 2) Compared with most approaches based on voxel grids, point clouds and meshes, our \nickname{} learns 3D representations without requiring 3D data for training; and 3) Compared with the existing implicit representation methods such as SDF \cite{Park2019}, SRNs \cite{Sitzmann2019} and NeRF \cite{Mildenhall2020}, our \nickname{} can represent diverse 3D contents from 2D views with strong generalization to novel geometries. In addition, the learned 3D representations carefully consider the general geometric patterns for every 3D spatial location, allowing the rendered views to be exceptionally realistic with fine-grained details. Figure \ref{fig:grf_intro} shows qualitative results of our \nickname{} which infers high-quality novel views for new objects of both seen and unseen categories.  Our key contributions are:
\begin{itemize}[leftmargin=*,topsep=1pt]
\itemsep-0.2em
    \item We propose a general radiance field to represent 3D structures and appearances from 2D images. It has strong generalization to novel geometries in a single forward pass.
    \item We integrate multi-view geometry and an attention mechanism to learn general geometric local patterns for each 3D query point along every query light ray. This allows the synthesized 2D views to be superior.
    \item We demonstrate significant improvement over baselines on large-scale datasets and provide intuition behind our design choices through extensive ablation studies. 
\end{itemize}

We note that some concurrent works such as pixelNeRF \cite{Yu2021}, IBRNet \cite{Wang2021}, SRF \cite{Chibane2021} and ShaRF \cite{Rematas2021} share the similar idea with GRF. The key difference is that we use geometry-aware attention module to combine the general 2D local features from multi-views, so that visual occlusions can be effectively addressed for better generalization.

\section{Related Work}
\label{sec:related}
\begin{figure*}[t]
\setlength{\belowcaptionskip}{ -8 pt}
\centering
   \includegraphics[width=1.\linewidth]{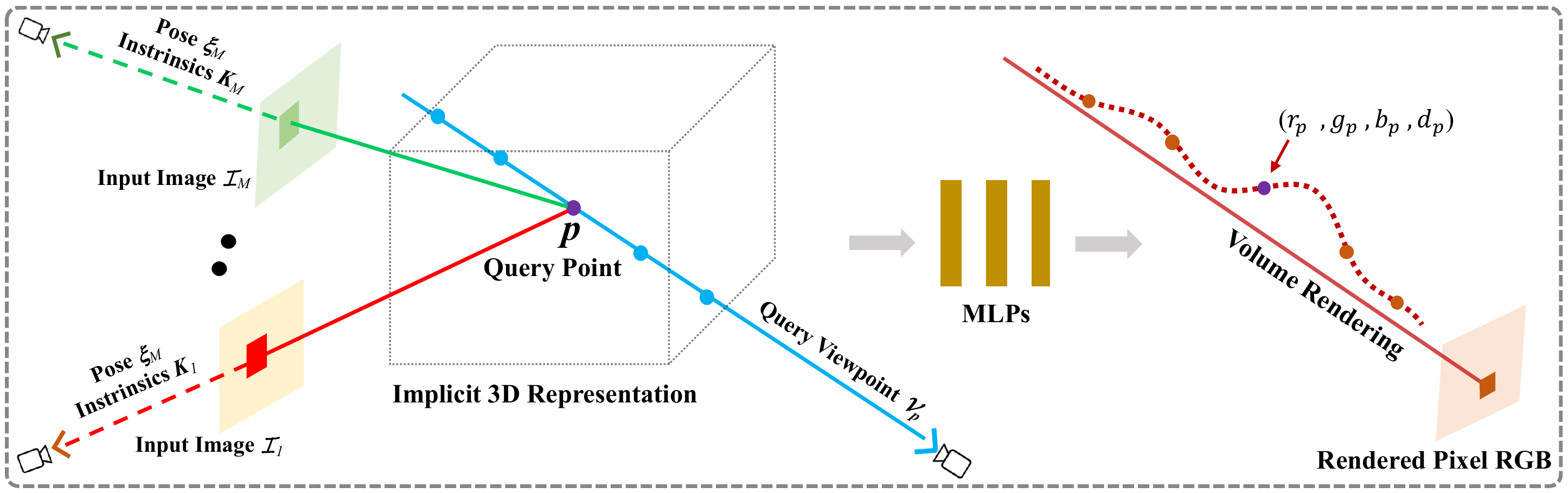}
\caption{Our \nickname{} projects each 3D point, $p$, to each of the $M$ input images, gathering per-pixel features from each view. These features are aggregated and fed to an MLP to infer $p$ with its color and volumetric density.}
\label{fig:overview}
\end{figure*}

\textbf{Classic Multi-view Geometry.}
Classic approaches to reconstruct 3D geometry from images mainly include SfM and SLAM systems such as Colmap \cite{Schonberger2016} and ORB-SLAM \cite{Mur-Artal2015}, which firstly extract and match hand-crafted geometric local features and then apply bundle adjustment for both shape and camera motion estimation \cite{Hartley2004}. Although they can recover visually satisfactory 3D models, the reconstructed shapes are usually sparse, discrete point clouds. In contrast, our \nickname{} learns to represent the continuous 3D structures from images. 

\textbf{Geometric Deep Learning.}
Impressive progress in recovering explicit 3D shapes from either single or multiple images has come from recent advances in deep neural nets such as voxel grid \cite{Chan2016,Song2017,Tulsiani2017c,Yang2018}, octree \cite{Riegler2017a,Christian2017}, point cloud \cite{Fan2017,Qi2016} and triangle mesh \cite{Wang2018f,Groueix2018,Gkioxari2019,Nash2020} approaches.
Although these methods can predict realistic 3D structures, they have two limitations. First, most of them require ground truth 3D labels to supervise the networks, resulting in the inability of the learned representations to generalize to novel real-world scenarios. Second, since the recovered 3D shapes are discrete, they are unable to preserve high-resolution geometric details. In contrast, our \nickname{} learns continuous 3D shape representations only from a set of 2D images with camera poses and intrinsics, which can be cheaply acquired, and also allow better generalization across new scenarios.

\textbf{Neural Implicit 3D Representations.}
The implicit representation of 3D shapes recently emerges as a promising direction to recover 3D geometries. It is initially formulated as level sets by optimizing neural nets which map \textit{xyz} locations to an occupancy field \cite{Mescheder2019,Saito2019} or a distance function \cite{Park2019}. The subsequent works \cite{Niemeyer2019,Sitzmann2019,Liu2019b,Atzmon2020,Atzmon2020a} introduce differentiable rendering functions, allowing 2D images or raw 3D point clouds to supervise the networks. Using neural radiance fields instead, the latest NeRF \cite{Mildenhall2020} and the succeeding NSVF \cite{Liu2020a}, NeRF-wild \cite{Martin-Brualla2020}, GRAF \cite{Schwarz2020} demonstrate impressive results to represent complex 3D environments. However, they do not take into account the local geometric patterns for spatial locations, thereby yielding less realistic rendered 2D images. Additionally, both NeRF and NeRF-wild can only represent a single scene, and are unable to generalize to novel scenarios. Uniquely, our \nickname{} maps any set of images to the corresponding 3D structure with geometric details.

\textbf{Novel View Synthesis and Neural Rendering.}
Novel view synthesis involves generating unseen views from multiple images. Existing methods usually learn a global embedding and then estimate a new image given a viewing angle, including GAN based methods \cite{Goodfellow2014,Radford2016}, variational auto-encoders \cite{Kingma2014}, autoregressive models \cite{Oord2016a}, and other generative frameworks \cite{Eslami2018}. Although photo-realistic single images can be generated, these methods tend to learn the manifold of 2D images, instead of exploiting the underlying 3D geometry for consistent multi-view synthesis. 

Neural rendering techniques \cite{Fried2020,Kato2020} have recently been investigated and integrated into 3D reconstruction pipelines, where there are no ground truth 3D data available but only 2D images for supervision. To render discrete voxel grids \cite{Yan2016,Rezende2016,Tulsiani2017}, point clouds \cite{Insafutdinov2018}, meshes \cite{Kato2017,Chen2019b,Liu2019e}, and implicit surfaces \cite{Liu2019b,Remelli2020}, most of these techniques are designed with differentiable and approximate functions, but sacrifice the sharpness of synthesized images. By contrast, our \nickname{} leverages the successful volume rendering of NeRF \cite{Mildenhall2020} which is naturally differentiable and can accurately render the RGB per light ray.

\section{\nickname{}}
\label{sec:grf}
\subsection{Overview}

Our \nickname{} models the 3D geometry and appearance as a neural network $f_\mathbf{W}$ where $\mathbf{W}$ represent learnable parameters. This network takes a set of $M$ images together with their camera poses and intrinsics $\{(\mathcal{I}_1, \boldsymbol{\xi}_1, \boldsymbol{K}_1) \cdots (\mathcal{I}_m, \boldsymbol{\xi}_m, \boldsymbol{K}_m) \cdots (\mathcal{I}_M, \boldsymbol{\xi}_M, \boldsymbol{K}_M) \}$, a query 3D point $p =\{ x_p, y_p, z_p\}$, and its query viewpoint $\mathcal{V}_p = \{x_p^v, y_p^v, z_p^v\}$ as input, and then predicts the RGB value $\{r_p, g_p, b_p\}$ and the volumetric density $d_p$ of that point $p$ observed from the viewpoint $\mathcal{V}_p$. Formally, it is defined as below:
\begin{align}
\setlength{\abovedisplayskip}{-0.pt}
\setlength{\belowdisplayskip}{-0.pt}
    (r_p, g_p, b_p, d_p) =& f_\mathbf{W}\Big(\big\{(\mathcal{I}_1, \boldsymbol{\xi}_1, \boldsymbol{K}_1), \cdots (\mathcal{I}_M, \boldsymbol{\xi}_M, \boldsymbol{K}_M)\big \}, \nonumber \\ 
    &\big\{x_p, y_p, z_p \big\}, \big\{x_p^v, y_p^v, z_p^v\big\} \Big)
\end{align}

The network is a general radiance field (\nickname{}) which parameterizes arbitrary 3D contents observed by the input $M$ images, returning both the appearance and geometry, when being queried at any location $p$ from any viewpoint $\mathcal{V}$ in 3D space. 

As illustrated in Figure \ref{fig:overview}, the proposed \nickname{} firstly extracts general features for each light ray through every pixel shown by the \textit{red} and \textit{green} square patches, and then reprojects those features back to the query 3D point $p$. After that, the corresponding RGB value and volumetric density $(r_p, g_p, b_p, d_p)$ are inferred from those features via MLPs, as shown by the \textit{purple} dot. By using volume rendering, multiple points on the same light ray are integrated, obtaining the rendered pixel RGB. 

Our network consists of four components: 1) A feature extractor for every 2D pixel; 2) A reprojector to transform 2D features to 3D space; 3) An aggregator to obtain general features for 3D points; and 4) The neural renderer NeRF \cite{Mildenhall2020} to infer the appearance and geometry for 3D points. 

\subsection{Extracting General Features for 2D Pixels}
\label{sec:pix_features}

\begin{wrapfigure}[9]{R}{0.25\textwidth}
\raisebox{0pt}[\dimexpr\height-.8\baselineskip\relax]{
\centering
\includegraphics[width=1\linewidth]{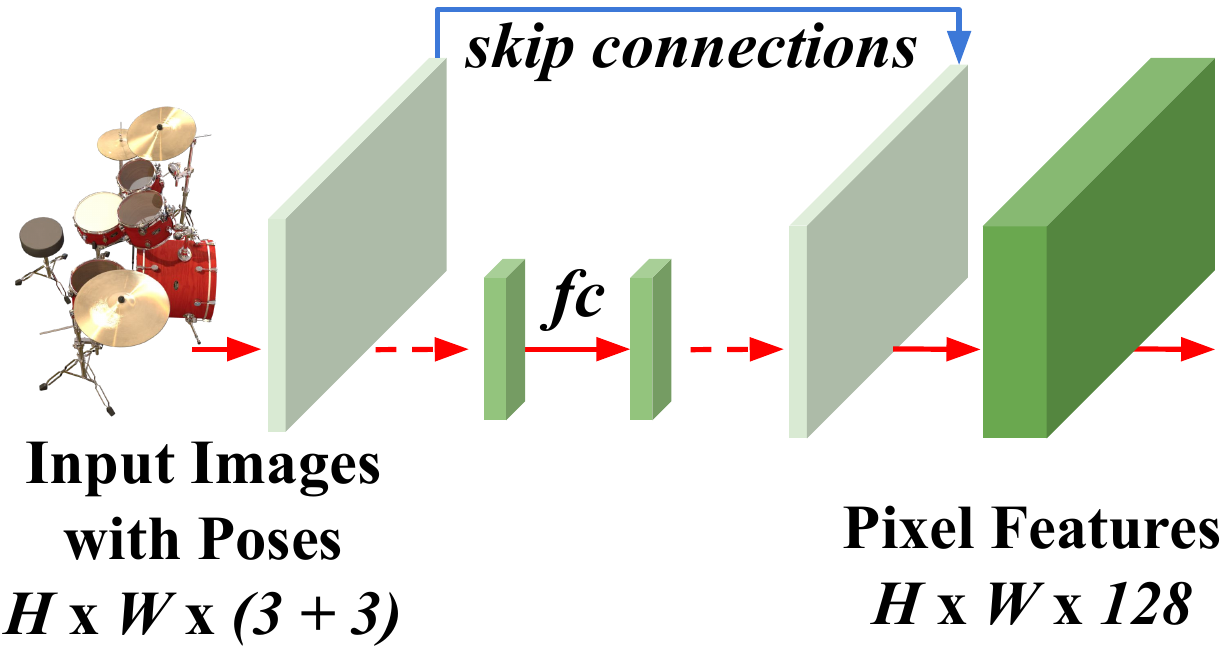}}
\caption{\small{Our CNN module extracts robust per-pixel features from each input view using skip connections.}}
\label{fig:pixel_features}
\end{wrapfigure}

Since each pixel of 2D images describes specific 3D points in space, this module is designed to extract the general features of each pixel, in order to learn the regional description and geometric patterns for each light ray. A naive approach is to directly use the raw \textit{rgb} values as the pixel features. However, this is sub-optimal because the raw \textit{rgb} values are sensitive to lighting conditions, environmental noise, etc. In order to learn more general and robust patterns for each pixel, we turn to use a more powerful encoder-decoder based convolutional neural network (CNN). As shown in Figure \ref{fig:pixel_features}, our CNN module is designed with the following two features:

\begin{itemize}[leftmargin=*]
\itemsep-0.1em
\item Instead of directly feeding raw RGB images into the CNN module, we stack (duplicate) the corresponding viewpoint, \textit{i.e.}, the $xyz$ location of the camera, to each pixel of the image. This allows the learned pixel features to be explicitly aware of its relative position in the 3D space. Note that, we empirically find that stacking the additional camera rotation and intrinsics to each pixel does not noticeably improve the performance.

\item We use skip connections between the encoder and decoder to preserve high frequency local features for each pixel, while optionally integrating a couple of fully connected (fc) layers in the middle of the CNN module to learn global features. The mixture of hierarchical features tends to to be more general and representative, effectively aiding the network in practice. 
\end{itemize}

\setlength{\columnsep}{10pt}

Details of the CNN module are presented in appendix and all input images share the same CNN module. Note that there are many ways to extract pixel features, but identifying an optimal CNN module is not in the scope of this paper.

\subsection{Reprojecting 2D Features to 3D Space}

\setlength{\columnsep}{10pt}
\begin{wrapfigure}[11]{R}{0.25\textwidth}
\raisebox{0pt}[\dimexpr\height-.5\baselineskip\relax]{
\centering
\includegraphics[width=1\linewidth]{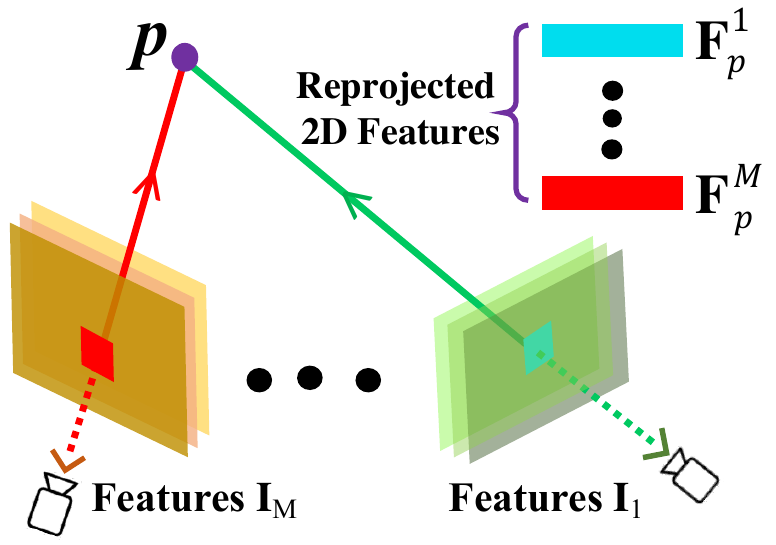}}
\caption{\small{Reprojecting pixel features back to a 3D point $p$.}}
\label{fig:grf_reproj}
\end{wrapfigure}

Considering that the extracted pixel features are a compact description of the light ray emitted from the camera center up to the 3D surface, we naturally reproject the pixel features back to the 3D space along the light ray. Since there are no depth scans paired with RGB images, it is impossible to determine which particular 3D surface point the pixel features belong to. In this module, we preliminarily regard the pixel features as the representation of every location along the light ray in 3D space. With this simple formulation, every 3D point can theoretically have a copy of its corresponding 2D pixel features from each 2D image. Formally, given a 3D point $p$, an observed 2D view $\mathcal{I}_m$ together with the camera pose $\boldsymbol{\xi}_m$ and the intrinsics $\boldsymbol{K}_m$, the corresponding 2D pixel features $\mathbf{F}_p^m$ are retrieved by the reprojection operation below:

\begin{equation}
\setlength{\abovedisplayskip}{-3.5pt}
    \mathbf{F}_p^m = \mathcal{P}\Big( \{\mathcal{I}_k, \boldsymbol{\xi}_m, \boldsymbol{K}_m\}, \{x_p, y_p, z_p\}, \mathbf{I}_m\Big)
\end{equation}
\textit{where} the function $\mathcal{P}()$ follows the principle of multi-view geometry \cite{Hartley2004} and $\mathbf{I}_m$ represents the image features extracted by the CNN module in Section \ref{sec:pix_features}. This reprojection is illustrated in Figure \ref{fig:grf_reproj}. However, since the pixels of 2D images are discrete and bounded within a certain spatial size, while the 3D points are continuous in the space, after the 3D point $p$ is projected to the plane of image $\mathcal{I}_m$, we apply two approximations to deal with the following issues.

\begin{itemize}[leftmargin=*]
\itemsep-0.1em
    \item If the point lies inside of the image, we simply select the nearest pixel and duplicate its features to the 3D point. Note that, more advanced techniques may be applied to address the discretization issue, such as bilinear interpolation or designing a kernel function.
    \item If the point lies outside of the image, we assign a zero vector to the 3D point, which means there is no information observed. In fact, we empirically find that the nearest interpretation can also achieve good performance, but it is only applicable for relatively small-scale structures.
\end{itemize}

Overall, the above simple reprojection operation explicitly retains the extracted 2D pixel features back to 3D space via the principle of geometry.

\subsection{Obtaining General Features for 3D Points}
For each query 3D point $p$, our \nickname{} retrieves a feature vector from each input image. However, given a set of input images, it is challenging to obtain a final feature vector for the point $p$, because:
\begin{itemize}[leftmargin=*]
\itemsep-0.1em
\item The total number of input images for each 3D scenario is variable and there is no order for images. Consequently, the retrieved feature vectors are also unordered with arbitrary size.  
\item Since there are no depth scans paired with the input RGBs, it is unable to decide which features are the true descriptions of the query point due to visual occlusions. Ideally, these features can be aware of the relative distance to the query point and then selected automatically.   
\end{itemize}

To tackle these critical issues, we formulate this problem as an attention aggregation process. In particular, as shown in Figure \ref{fig:grf_aggregation}, given the query 3D point $p$, its query viewpoint $\mathcal{V}_p$, and the set of retrieved pixel features $\{ \mathbf{F}_p^1 \cdots \mathbf{F}_p^m \cdots \mathbf{F}_p^M \}$:
\begin{itemize}[leftmargin=*]
\itemsep-0.1em
\item For each retrieved feature vector $\mathbf{F}_p^m$, we firstly use shared MLPs to integrate the information of query point $p$, generating a new feature vector $\mathbf{\hat{F}}_p^m$ which is aware of the relative distance to the query point $p$. Formally, it is defined as:
\begin{equation*}
\setlength{\abovedisplayskip}{4pt}
\setlength{\belowdisplayskip}{4pt}
    \mathbf{\hat{F}}_p^m = MLPs(\mathbf{F}_p^m \oplus [x_p, y_p, z_p]), ( \oplus \mbox{ is concatenation})
\end{equation*}
\item After obtaining the new set of position-aware features $\{\mathbf{\hat{F}}_p^1 \cdots \mathbf{\hat{F}}_p^m \cdots \mathbf{\hat{F}}_p^M \}$, we use the existing attention aggregation methods such as AttSets \cite{Yang2020} and Slot Attention \cite{Locatello2020} to compute a unique feature vector $\mathbf{\bar{F}}_p$ for the query 3D point $p$. Basically, the attention mechanism learns a unique weight for all input features and then aggregates them together. According to the theoretical analysis in \cite{Yang2020} and \cite{Locatello2020}, the selected attention mechanisms are permutation invariant with regard to the input set of feature vectors and can process an arbitrary number of elements. Formally, it is defined as:
\begin{figure}[t]
\setlength{\abovecaptionskip}{ 4. pt}
\setlength{\belowcaptionskip}{ -12 pt}
\centering
   \includegraphics[width=1.\linewidth]{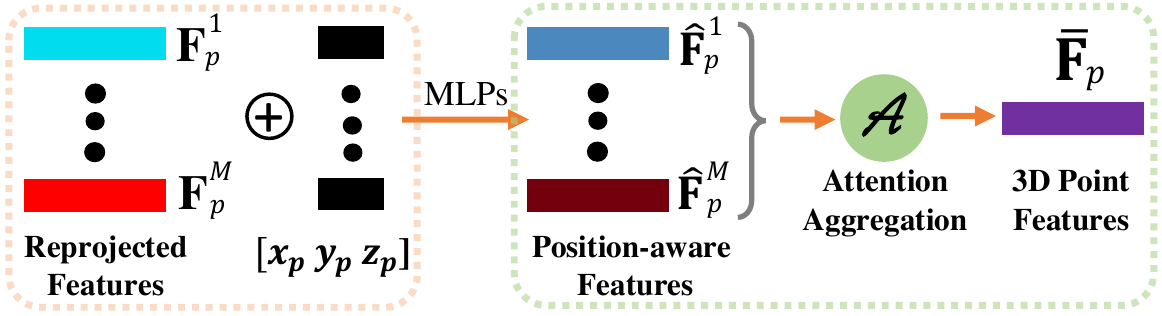}
\caption{Pixel features are aggregared over all input views for each 3D point.}
\label{fig:grf_aggregation}
\end{figure}
\begin{equation*}
    \mathbf{\bar{F}}_p = \mathcal{A}\big(\mathbf{\hat{F}}_p^1 \cdots \mathbf{\hat{F}}_p^m \cdots \mathbf{\hat{F}}_p^M \big), ( \mathcal{A} \mbox{ is an attention function})
\end{equation*}
\end{itemize}
\vspace{-0.45\baselineskip}

Above all, for every query 3D point $p$, its final features $\mathbf{\bar{F}}_p$ explicitly preserve the general geometric patterns retrieved from the input 2D observations and are also made aware of the query point location in 3D space. This allows the 3D point features $\mathbf{\bar{F}}_p$ to be general and representative for its own geometry and appearance.

\subsection{Rendering 3D Features via NeRF}
\label{sec:nerf}
\begin{figure}[ht]
\setlength{\abovecaptionskip}{ 4. pt}
\centering
   \includegraphics[width=1.\linewidth]{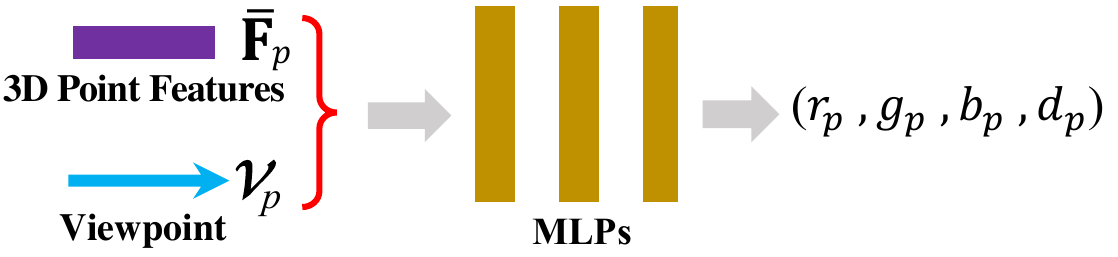}
\caption{The aggregated point features and viewing direction are concatenated as input to an MLP to predict color and density for every point.}
\label{fig:grf_rendererNerf}
\end{figure}

For any query 3D point $p$, we feed its features $\mathbf{\bar{F}}_p$ and query viewpoint $\mathcal{V}_p$, \textit{i.e.}$(x_p^v, y_p^v, z_p^v)$, into MLPs, and then predict its RGB values $\{r_p, g_p, b_p\}$ and volumetric density $d_p$. In fact, these MLPs forms a general radiance field. As illustrated in Figure \ref{fig:grf_rendererNerf}, we exactly follow the MLPs designed in NeRF \cite{Mildenhall2020}. Note that, the only difference is that our \nickname{} uses the point features $\mathbf{\bar{F}}_p$ as input, while the original NeRF uses the point position $xyz$.

An image pixel RGB can be rendered from the radiance field by casting a ray from the camera center towards the 3D space using volume rendering equations \cite{Kajiya1984} below:
\begin{equation*}
    rgb = \rint_0^{+\infty} T(t)\boldsymbol{c}(t)\sigma(t)dt,\quad T(t) = exp(-\rint_0^t\sigma(s)ds)
\end{equation*}
\textit{where} $\boldsymbol{c}(t)$ and $\sigma(t)$ are the color and volume density at point $t$ on the ray. The above integrals are estimated with numerical quadrature as in NeRF \cite{Mildenhall2020} with a hierarchical sampling strategy. We strictly follow this method to estimate the color for each ray. Thus, our \nickname{} can directly synthesize novel 2D images by querying points along light rays. This allows the entire network to be trainable only with a set of 2D images, without requiring 3D data. 

\subsection{Implementation}
The above four modules are connected and trained end-to-end. Details of the CNN module and the attention module for different experiments are presented in appendix. All the designs of neural rendering strictly follow NeRF \cite{Mildenhall2020}. In our network, all 3D locations and RGB values are processed by the positional encoding proposed by NeRF. The L2 loss between rendered RGBs and the ground truth is used to optimize the whole network.

\section{Experiments}\label{sec:exp}
\subsection{Generalization to Unseen Objects}
\label{sec:exp_novelObj}
\begin{figure*}[ht]
\setlength{\abovecaptionskip}{ 1 pt}
\setlength{\belowcaptionskip}{ -1 pt}
  \centering
  \includegraphics[width=1\linewidth]{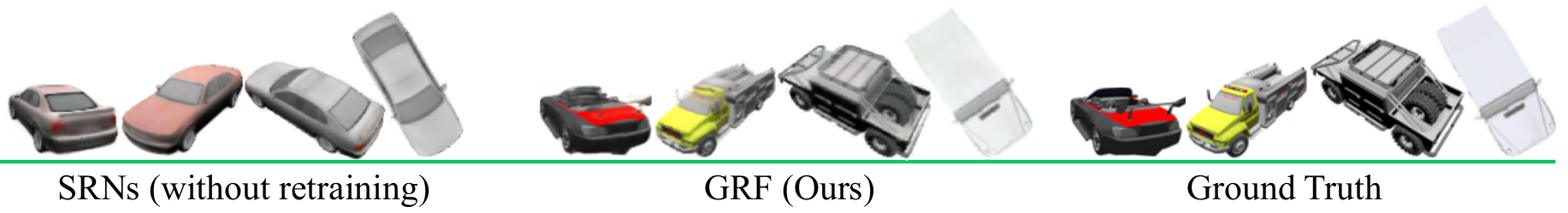}
  \caption{Qualitative results of SRNs (without retraining) and our \nickname{} for novel view synthesis of unseen cars. SRNs fails to recover faithful shapes for unseen cars without retraining.}
  \label{fig:shapenet_srns_fail}
\end{figure*}

Following the experimental settings of SRNs \cite{Sitzmann2019}, we firstly evaluate the novel view synthesis of our \nickname{} on the chair and car classes of ShapeNetv2. Particularly, the chair has 4612 objects for training, 662 for validation and 1317 for testing, while the car has 2151 objects for training, 352 for validation and 704 for testing. Each training object has randomly sampled 50 images with a resolution of 128$\times$128. We train two separate models on chairs and cars and then conduct the following two groups of experiments.
\begin{itemize}[leftmargin=*]
\itemsep-0.1em
\item Group 1: Novel-view synthesis of unseen objects in the testing split of the same category. The trained two models are tested on novel objects of the same category. During testing, the model is fed with 2 novel views of each novel object, inferring 251 novel views for evaluation. 
\item Group 2: Similar to Group 1, but only 1 novel view is fed into the model for novel view synthesis. 
\end{itemize}

Table \ref{table:exp_novelObj} compares the quantitative results of our \nickname{} and four baselines. Note that, the recent NeRF and NSVF are not scalable to learn large number of scenes simultaneously because each scene is encoded into the network parameters.

\setlength\tabcolsep{1.5pt}
\begin{table}[t!]
	\centering
	\ra{1.}
	\caption{Comparison of the average PSNR (in dB) and SSIM of reconstructed images in the ShapeNetv2 dataset. The higher the scores, the better the synthesized novel views. \textbf{Note that, our \nickname{} solves a harder problem than SRNs, because \nickname{} infers the novel object representation in a single forward pass, while SRNs cannot.}}
	\small
	\begin{tabularx}{0.47\textwidth}{@{}rccrcc@{}}
		\toprule
		& \multicolumn{2}{c}{2 Images (Group 1)} 
		& \phantom{} 
		& \multicolumn{2}{c}{1 Image (Group 2)} \\
		\cmidrule{2-3} \cmidrule{5-6}
		& Chairs 	& Cars 	
		&& Chairs 	& Cars \\ 
		\midrule
		TCO	\cite{Tatarchenko2015}	
		& $21.33$ / $0.88$ & $18.41$ / $0.80$
		&& $21.27$ / $0.88$ & $18.15$ / $0.79$ \\
		WRL	\cite{Worrall2017}
		& $22.28$ / $0.90$	& $17.20$ / $0.78$	
		&& $22.11$ / $0.90$	& $16.89$ / $0.77$ \\
		dGQN \cite{Eslami2018} 		
		& $22.36$ / $0.89$	& $18.79$ / $0.79$ 	
		&& $21.59$ / $0.87$	& $18.19$ / $0.78$	\\
		SRNs \cite{Sitzmann2019}							
		& $\textbf{24.48}$ / $\textbf{0.92}$	& $\textbf{22.94}$ / $\textbf{0.88}$	
		&& $\textbf{22.89}$ / $\textbf{0.91}$	& $\textbf{20.72}$ / $\textbf{0.85}$ \\
		\textbf{\nickname{}(Ours)} 		
		& $\underline{22.65}$ / $\underline{0.88}$	& $\underline{22.34}$ / $\underline{0.86}$ 	
		&& $21.25$ / $0.86$	& $\underline{20.33}$ / $\underline{0.82}$	\\
		\bottomrule
	\end{tabularx}
	\label{table:exp_novelObj}
\end{table}

\textbf{Analysis.} Our \nickname{} achieves comparable performance with SRNs on the car category for novel view synthesis in both Group 1\&2. Note that, our \nickname{} solves a much harder problem than SRNs. In particular, our network directly infers the unseen object representation in a single forward pass, while SRNs needs to be retrained on all new objects to optimize the latent code. As a result, the experiments of Group 1\&2 are significantly in favour of SRNs. 

To demonstrate the advantage of \nickname{} over SRNs, we directly evaluate the trained SRNs model on unseen objects (of the same category) without retraining. For comparison, we also directly evaluate the trained GRF model on the same novel objects. Figure \ref{fig:shapenet_srns_fail} shows the qualitative results. It can be seen that if not retrained, SRNs completely fails to reconstruct unseen car instances, but randomly generates similar cars from learned prior knowledge, primarily because its latent code has not been updated from the unseen objects. In contrast, by learning pixel local patterns, our \nickname{} generalizes well to novel objects directly.

\subsection{Generalization to Unseen Categories}
\label{sec:exp_novelCat}

We further evaluate the generalization capability of our \nickname{} across unseen object categories on the ShapeNet dataset rendered by DISN \cite{Xu2019d}. In particular, we train a single model of our network on 6 categories \{chair, bench, car, airplane, table, speaker\}, and then directly test it on the remaining 7 unseen categories \{cabinet, display, lamp, phone, rifle, sofa, watercraft\}. For each training category, 1000 objects are randomly selected, while for each testing category, 200 objects are randomly selected. All objects have 36 rendered images with 224$\times$224 pixels. For comparison, we also train a single model for SRNs on 6 categories. During testing, we carefully retrain SRNs on all objects of the 7 categories. The following two groups of experiments for view synthesis are conducted.
\begin{itemize}[leftmargin=*]
\itemsep-0.1em
\item Group 1: Both our \nickname{} model and the retrained SRNs model are fed with 2 views of each object from unseen categories, inferring the total 36 views for evaluation. 
\item Group 2: Similar to Group 1, but 6 views are fed into the two models for novel view synthesis. 
\end{itemize}

Table \ref{table:exp_novelCat} shows the quantitative results of \nickname{} and SRNs. We also include the scores for novel objects of trained categories for comparison. It can be seen that our \nickname{} maintains the similar performance when evaluated on unseen categories, demonstrating the strong generalization capability. Notably, thanks to our attentive aggregation module, given more input images (6 \textit{vs} 2), the overall performance and generalization of \nickname{} increases significantly, while SRNs improves marginally even though it is retrained.  Figure \ref{fig:grf_intro} shows qualitative results. More results are in appendix.

\setlength\tabcolsep{1.5pt}
\begin{table}[t!]
	\centering
	\ra{1.}
	\caption{Comparison of the average PSNR and SSIM of reconstructed images by our \nickname{} and SRNs \cite{Sitzmann2019}. The higher the scores, the better the synthesized novel views. \textbf{Note that, the SRNs is retrained on the new 7 classes.}}
	\small
	\begin{tabularx}{0.47\textwidth}{@{}lccrcc@{}}
		\toprule
		& \multicolumn{2}{c}{Unseen 7 Classes} 
		& \phantom{abc} 
		& \multicolumn{2}{c}{Seen 6 Classes \scriptsize{(New Objects)}} \\
		\cmidrule{2-3} \cmidrule{5-6}
		& Group 1 	& Group 2
		&& Group 1	& Group 2 \\ 
		& (2 Images)  & (6 Images)
		&& (2 Images) & (6 Images) \\
		\midrule
		SRNs							
		& $24.16$ / $0.90$	& $25.76$ / $0.91$ 	
		&& $\textbf{25.74}$ / $\textbf{0.91}$	& $26.79$ / $0.92$	\\
		\textbf{Ours} 		
		& $\textbf{24.68}$ / $\textbf{0.90}$	& $\textbf{29.37}$ / $\textbf{0.95}$	
		&& $25.63$ / $0.90$	& $\textbf{29.57}$ / $\textbf{0.95}$ \\
		\bottomrule
	\end{tabularx}
	\label{table:exp_novelCat}
\end{table}

\subsection{Generalization to Unseen Scenes}
\label{sec:exp_genScene}
We further evaluate the generalization of \nickname{} on a more complex dataset Synthetic-NeRF \cite{Mildenhall2020}. It consists of pathtraced images of 8 synthetic scenes with complicated geometry and realistic materials. Each scene has 100 views for training and 200 novel views for testing. Each image has 800 $\times$ 800 pixels. We train a single model on randomly selected 4 scenes, i.e., Chair, Mic, Ship, and Hotdog, and then conduct the following two groups of experiments.
\begin{itemize}[leftmargin=*]
\itemsep -0.1em
\item Group 1 (Without Finetuning): The trained model is directly tested on the remaining 4 novel scenes, i.e., Drums, Lego, Materials, and Ficus. Basically, this experiment is to evaluate whether the learned features can truly generalize to new scenarios. This is extremely challenging because there are only 4 training scenes and the overall shapes of the 4 novel scenes are dramatically different.
\item Group 2 (With Finetuning): The trained model is further finetuned on each of the four novel scenes with $100$, 1k, and 10k iterations separately. In total, we obtain (3 models/scene $\times$ 4 scenes = 12 new models). For comparison, we also train NeRF on each of the four novel scenes with $100$, 1k, 10k iterations from scratch. This group of experiments evaluates how the initially-learned features of our \nickname{} can be transferred to novel scenes.
\end{itemize}

\begin{figure*}[th]
\setlength{\abovecaptionskip}{ 1 pt}
\setlength{\belowcaptionskip}{ -1 pt}
\centering
   \includegraphics[width=1\linewidth]{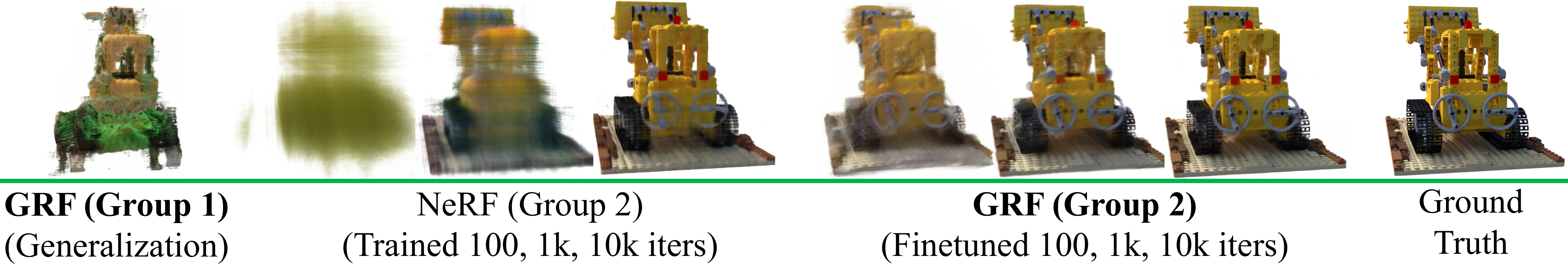}
\caption{Qualitative results of our \nickname{} for novel scene generalization. }
\label{fig:synNeRF_genalization}
\end{figure*}

\setlength\tabcolsep{1.5pt}
\begin{table}[t!]
	\centering
	\tabcolsep=0.16cm
     \caption{The average scores of PSNR, SSIM and LPIPS for \nickname{} and NeRF on four novel scenes of Synthetic-NeRF.}
      \begin{tabular}{lcccc} \toprule
         & PSNR$\uparrow$ & SSIM$\uparrow$ & LPIPS$\downarrow$ \\ \midrule
         \textbf{\nickname{} (Group 1)} & 13.62 & 0.763 & 0.246	 \\ \midrule
         NeRF (Group 2, 100 iters) &  15.15 & 0.752 & 0.359 \\
         NeRF (Group 2, 1k iters) &  19.81 & 0.809 & 0.228 \\
         NeRF (Group 2, 10k iters) &  23.35 & 0.875 & 0.137 \\ \midrule
         \textbf{\nickname{} (Group 2, 100 iters)} & 19.69 & 0.835 & 0.169 \\
         \textbf{\nickname{} (Group 2, 1k iters)} & 22.00 & 0.876 & 0.128 \\
         \textbf{\nickname{} (Group 2, 10k iters)} & 25.10 & 0.916 & 0.089 \\ \bottomrule
      \end{tabular}
	\label{table:synNeRF_genalization}
\end{table}

\textbf{Analysis.} Table \ref{table:synNeRF_genalization} compares the quantitative results and Figure \ref{fig:synNeRF_genalization} shows the qualitative results. We can see that: 1) In Group 1, our \nickname{} can indeed generalize to novel scenes with complex geometries, demonstrating the effectiveness of learning point local features in \nickname{}. As shown in the first image of Figure \ref{fig:synNeRF_genalization}, the overall shape and appearance of Lego can be satisfactorily recovered, though it has never been seen before. 2) In Group 2, our \nickname{} can quickly learn high-quality scene representations given a small number of training iterations, thanks to the initially learned general features. Compared with the NeRF trained from scratch, the learned \nickname{} significantly speed up novel scene learning, achieving much better results given the same training iterations of new scenes. Additionally, we conduct similar generalization experiments on the real-world dataset \cite{Choi2016}. More results are in appendix.

\subsection{Pushing the Boundaries of Single Scenes}
\label{sec:exp_singleScene}
\begin{figure*}[hbt]
\setlength{\abovecaptionskip}{ -0.1 pt}
\setlength{\belowcaptionskip}{ -2 pt}
\centering
   \includegraphics[width=0.99\linewidth]{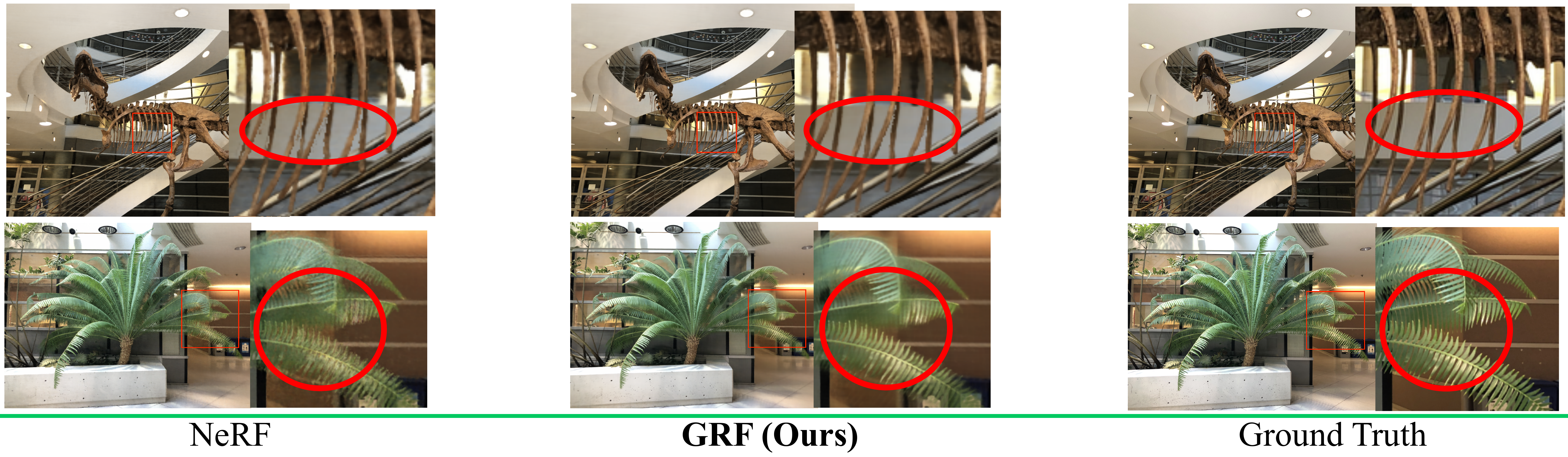}
\caption{Qualitative results on real-world scenes. Our \nickname{} can generate more realistic images than NeRF.}
\label{fig:syn_llff}
\end{figure*}

In addition to the generalization of our \nickname{} for unseen objects and scenes, the learned pixel features are expected to significantly improve the quality of rendered images for single scenes. To validate this, we conduct experiments on complex real-world scenes captured by cellphones. There are 8 scenes, 5 from LLFF \cite{Mildenhall2019} and 3 from NeRF. Each scene has 20 to 62 images, 1/8 of which for testing. All images have  1008 $\times$ 756 pixels. In particular, we train a single model for each real-world scene, following the same experimental settings of NeRF. Table \ref{table:exp_singlScene} compares the quantitative results. Note that, the recent NSVF \cite{Liu2020a} is unable to process these forward-facing scenes because the predefined voxels cannot represent the unbounded 3D space. 

\textbf{Analysis.} Our method surpasses the state of the art NeRF \cite{Mildenhall2020} by large margins, especially over the SSIM and LPIPS metrics. Compared with PSNR which only measures the average per-pixel accuracy, the metrics SSIM and LPIPS favor high quality of photorealism, highlighting the superiority of our \nickname{} to generate truly realistic images. Figure \ref{fig:syn_llff} shows the qualitative results. As highlighted by the red circles, our \nickname{} can generate fine-grained geometries in pixel level, while NeRF produces many artifacts. This demonstrates our \nickname{} indeed learns precise pixel features from the 2D images for 3D representation and rendering.

\setlength\tabcolsep{1.5pt}
\begin{table}[t!]
	\centering
	\tabcolsep=0.42cm
    \caption{The average scores of PSNR, SSIM and LPIPS in the challenging real-world dataset for single-scene learning.}
      \begin{tabular}{rcccc}
        \toprule
          & PSNR$\uparrow$ & SSIM$\uparrow$ & LPIPS$\downarrow$ \\
         \midrule
         SRNs \cite{Sitzmann2019} & 22.84 & 0.668 & 0.378 \\
         LLFF \cite{Mildenhall2019} & 24.13 & 0.798 & 0.212 \\
         NeRF \cite{Mildenhall2020}& 26.50 & 0.811 & 0.250\\
         \textbf{\nickname{}(Ours)} &\textbf{26.64} & \textbf{0.837} & \textbf{0.178} \\
         \bottomrule
      \end{tabular}
	\label{table:exp_singlScene}
\end{table}

\subsection{Ablation Study}
\label{sec:exp_ablation}

To evaluate the effectiveness of the key components of our \nickname{}, we conduct 3 groups of ablation experiments on the car category in ShapeNetv2 dataset. In particular, we train on the entire training split then randomly select 500 objects from the training split for novel view synthesis. During testing, the model infers 50 novel views for each of these 500 objects for evaluation.
\begin{itemize}[leftmargin=*]
\itemsep-0.1em
\item Group 1: The viewpoints of the input images are removed from the CNN module. The CNN module is not explicitly aware of the relative position of the pixel features.
\item Group 2: The decoder of the CNN module is removed and each input image is encoded as a global feature vector. For any 3D query point which is rightly projected into the image boundary, that global feature vector is retrieved and reprojected to the query point. Fundamentally, this modified CNN module can be regarded as a hyper-network that learns a conditional embedding from input images and then feed it into NeRF, but it sacrifices the precise pixel local features.
\item Group 3: The advanced attention module is replaced by max-pooling to aggregate the pixel features, aiming to investigate how the visual occlusion can be better addressed using soft attention.
\item Group 4: The full model is trained and tested with the same settings for comparison.
\end{itemize}

\textbf{Analysis.} Table \ref{table:ablation} compares the performance of all ablated models. We can see that: 1) The greatest impact is caused by the removal of image viewpoints from the CNN module and the lack of local pixel features to represent 3D points. It highlights that obtaining the position-aware and precise pixel features is crucial for 3D representation from 2D images, while learning a simple hyper-network for NeRF cannot achieve comparable performance. 2) Using max-pooling to select the reprojected pixel features is sub-optimal to address the visual occlusions.

\setlength\tabcolsep{1.5pt}
\begin{table}[t!]
	\centering
	\tabcolsep=0.25cm
    \caption{The average scores of PSNR and SSIM for ablated \nickname{} in the subset of car category.}
      \begin{tabular}{lccc}
        \toprule
          & PSNR$\uparrow$ & SSIM$\uparrow$  \\
         \midrule
         (1) Remove Input Viewpoints & 20.13 & 0.807  \\
         (2) Remove Pixel Local Features & 20.23 & 0.818  \\
         (3) Replace Attention by Maxpool & 24.88 & 0.914\\
         \textbf{(4) The Full Model} &\textbf{27.16} & \textbf{0.942} \\
         \bottomrule
      \end{tabular}
	\label{table:ablation}
\end{table}

\subsection{Analysis of Attention Mechanism}
\label{sec:exp_attention}

\begin{figure}[t]
\setlength{\belowcaptionskip}{ -2 pt}
\centering
   \includegraphics[width=0.99\linewidth]{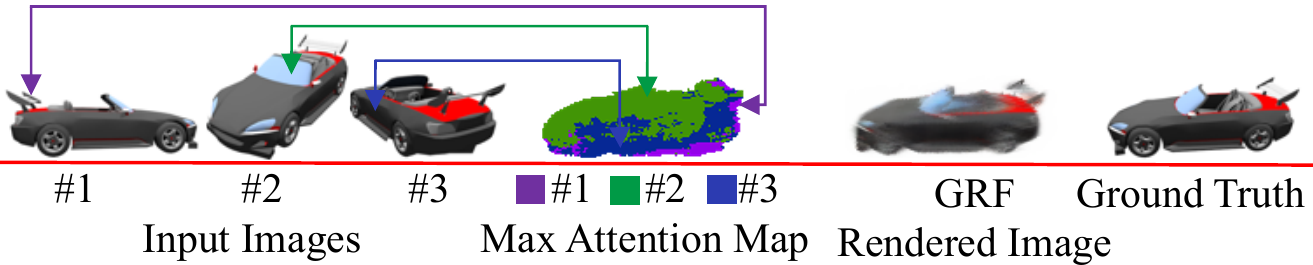}
\caption{The attention mechanism learns to select the useful pixel information for inferring a novel view.}
\label{fig:exp_attention}
\end{figure}

The attention mechanism in our \nickname{} aims to automatically select the correct pixel patch from multiple pixel patches where the light rays intersect at the same query 3D point in space. In order to investigate how the attention mechanism learns to select the useful information, as shown in Figure \ref{fig:exp_attention}, we feed three images (\#1,\#2,\#3) of an unseen car into our GRF model which is well-trained on ShapeNet car category, and then render a new image (i.e., the 5th image in Figure \ref{fig:exp_attention}). For each pixel of the rendered image, we retrieve the input image pixel that has the highest attention score, obtaining a Max Attention Map (the 4th image in Figure \ref{fig:exp_attention}). Specifically, the rendered pixels with purple color correspond to the input image \#1, the green pixels correspond to the input image \#2, while the blue pixels correspond to the input image \#3. Experiment details are in appendix.

It can be seen that, when inferring a new image, the attention module of our GRF focuses on the most informative pixel patch from the multiple input pixel patches. In addition, it is able to truly deal with the visual occlusion. For example, when inferring the windshield of the car, the attention module focuses on the input image \#2 where the windshield is visible, while ignoring the image \#1 and \#3 where the windshield is self-occluded.

\section{Conclusion}
Our proposed method models 3D geometries as a general radiance field. We have demonstrated that our \nickname{} can learn general and robust 3D point features from a set of sparse 2D observations by using the principle of multi-view geometry to precisely map 2D pixel features back to 3D space and by leveraging attention mechanisms to implicitly address visual occlusions. In doing so, our \nickname{} can synthesize truly realistic novel views. However, there are still limitations that may lead to future work: 1) more advanced CNN modules can be designed to learn better pixel features; and 2) depth scans can be integrated into the network to explicitly address visual occlusions. \\

\noindent\textbf{Acknowledgments:} This work was partially supported by HK PolyU (UGC) under Project P0034792.

\clearpage
{\small
\bibliographystyle{ieee_fullname}
\bibliography{references}
}

\clearpage
\onecolumn
\appendix
\section{Appendix}
\subsection{Details of Network Architecture}
\begin{table}[h]
\centering
\caption{The CNN Module for experiments on the ShapeNetv2 dataset in Sections \ref{sec:exp_novelObj} and \ref{sec:exp_novelCat}}
\begin{tabular}{lcccc}
\toprule
Type & Size/Channels & Activation & Stride \\
\midrule
Input: embedding of RGB and viewpoint & - & - & - \\ 
L1: Conv $7\times 7$ & $64$ & ReLU & $2$\\
L2: Conv $3\times 3$ & $128$ & ReLU & $2$\\
L3: Conv $3\times 3$ & $256$ & ReLU & $2$\\
L4: Conv $3\times 3$ & $512$ & ReLU & $2$\\
L5: Conv $4\times 4$ & $128$ & ReLU & $4$\\
L6: Flatten / Tile  & - & - & - \\
L7: Concat (L6, L4)  & - & - & - \\
L8: Dilated Conv $3\times3$ & $256$ & ReLU & 4\\
L8: Concat (L8, L3) & - & - &- \\
L9: Dilated Conv $3\times3$ & $128$ & ReLU & 8\\
L9: Concat (L9, L2) & - & - &- \\
L10: Dilated Conv $3\times3$ & $64$ & ReLU & 16 \\
L10: Concat (L10, L1) & - & - &- \\
L11: Dilated Conv $3\times3$ & $128$ & ReLU & 32 \\
\bottomrule
\end{tabular}
\label{table:cnn_shapenetv2}
\end{table}

\begin{table}[h]
\centering
\caption{The CNN Module for experiments on the Synthetic-NeRF dataset in Section \ref{sec:exp_genScene}. The average pooling is added to aggressively downsample the feature maps.}
\begin{tabular}{lcccc}
\toprule
Type & Size/Channels & Activation & Stride \\
\midrule
Input: embedding of RGB and viewpoint & - & - & - \\
L1: Conv $7\times 7$ & $64$ & ReLU & $2$\\
L2: Conv $3\times 3$ & $128$ & ReLU & $2$\\
L3: Conv $3\times 3$ & $256$ & ReLU & $2$\\
L4: Conv $3\times 3$ & $512$ & ReLU & $2$\\
L4: AveragePooling $5\times5$ & - & - & - \\
L5: Conv $5\times 5$ & $128$ & ReLU & $5$\\
L6: Flatten / Tile  & - & - & - \\
L7: Concat (L6, L4)  & - & - & - \\
L8: Deconv $3\times3$ & $256$ & ReLU & 2\\
L8: Concat (L8, L3) & - & - &- \\
L9: Deonv $3\times3$ & $128$ & ReLU & 2\\
L9: Concat (L9, L2) & - & - &- \\
L10: Deconv $3\times3$ & $64$ & ReLU & 2 \\
L10: Concat (L10, L1) & - & - &- \\
L11: Deconv $3\times3$ & $128$ & ReLU & 2 \\
\bottomrule
\end{tabular}
\label{table:cnn_synNeRF}
\end{table}

\begin{table}[htp!]
\centering
\caption{The CNN Module for experiments on the real-world dataset (LLFF) in Section \ref{sec:exp_singleScene}. The average pooling is added to aggressively downsample the feature maps.}
\begin{tabular}{lcccc}
\toprule
Type & Size/Channels & Activation & Stride \\
\midrule
Input: embedding of RGB and viewpoint& - & - & - \\
L1: Conv $7\times 7$ & $64$ & ReLU & $2$\\
L2: Conv $3\times 3$ & $128$ & ReLU & $2$\\
L3: Conv $3\times 3$ & $256$ & ReLU & $2$\\
L4: Conv $3\times 3$ & $512$ & ReLU & $2$\\
L4: AveragePooling $8\times8$ & - & - & - \\
L5: Conv $4\times 4$ & $128$ & ReLU & $4$\\
L6: Flatten / Tile  & - & - & - \\
L7: Concat (L6, L4)  & - & - & - \\
L8: Deconv $3\times3$ & $256$ & ReLU & 2\\
L8: Concat (L8, L3) & - & - &- \\
L9: Deconv $3\times3$ & $128$ & ReLU & 2\\
L9: Concat (L9, L2) & - & - &- \\
L10: Deconv $3\times3$ & $64$ & ReLU & 2 \\
L10: Concat (L10, L1) & - & - &- \\
L11: Deconv $3\times3$ & $128$ & ReLU & 2 \\
\bottomrule
\end{tabular}
\label{table:cnn_realworld}
\end{table}

\begin{table}[ht]
\centering
\caption{The Attention Module, AttSets \cite{Yang2020}, for experiments on the ShapeNetv2 Dataset in Sections \ref{sec:exp_novelObj} and \ref{sec:exp_novelCat}. The simple AttSets is computationally efficient and we choose it to train the large-scale ShapeNetv2 dataset.}
\begin{tabular}{lcccc}
\toprule
Type & Size/Channels & Activation  \\
\midrule
Input: Concat($K\times128$, embedding of viewpoint) & - & -  \\
L1: fc & $256$ & ReLU  \\
L2: fc & $256$ & ReLU \\
L3: fc & $256$ & ReLU \\
L4: fc & $512$ & ReLU \\
L5: fc & $512$ & ReLU \\
L6: softmax(L5) & - & - \\
L7: sum(L6*L5, axis=-2) & - & - \\
L8: fc & $512$ & ReLU \\
\bottomrule
\end{tabular}
\label{table:attention_shapenetv2}
\end{table}
\newpage
We use Slot Attention as the pixel feature aggregation module for experiments on the Synthetic-NeRF and the real-world dataset in Sections \ref{sec:exp_genScene} and \ref{sec:exp_singleScene}. In particular, we use two slots, two iterations, and the hidden size is 128. The final output two slots are flattened and a 256 dimensional vector is obtained. 

For details of Slot Attention refer to the paper \cite{Locatello2020}.
Details of the neural rendering layers and the volume rendering can be found in NeRF \cite{Mildenhall2020}. We set the positional embedding length $L=5$ for all inputs to the CNN module, except the rotation, which we convert to quaternion and embed at $L=4$.

During training, we feed the models between 2 and 6 views of each geometry at each gradient step. We set the learning rate for the ShapeNetv2 models at 1e-4. We set the learning rate for leaves and orchids in the real-world dataset at 7e-5, and for the rest, we use 1e-4. For Synthetic-NeRF dataset, we use a learning rate of 1e-4. We use the Adam optimizer for all models, and train for 200k-300k iterations. At each gradient step, we take 1000 rays for ShapeNetv2 with 32 coarse samples and 64 fine samples, and 800 rays for the real-world and Synthetic-NeRF datasets with 64 coarse samples and 192 fine samples. We train each model on a single Nvidia-V100 GPU with 32GB VRAM. 

During testing on the ShapeNetv2 dataset in Section \ref{sec:exp_novelObj}, we feed the model the 4 closest views by cosine similarity to the desired novel view.

\clearpage

\def\arraystretch{0.85} 
\subsection{Details of Experimental Results on the Synthetic-NeRF Dataset in Section \ref{sec:exp_genScene}}
\begin{table*}[ht]
\tabcolsep=0.cm
\small
\centering
\caption{The PSNR, SSIM and LPIPS scores of our \nickname{} simultaneously trained on 4 scenes of the Synthetic-NeRF dataset for multi-scene learning in Section \ref{sec:exp_genScene}. The scores of SRNs, NeRF and NSVF trained on single scenes are included for comparison. 
\label{table:synNeRF_4scenes}}
\begin{tabular*}{1.\textwidth}{@{\extracolsep{\fill}}lcccc}
\toprule
 &  Chair  & Mic & Ship & Hotdog \\
\midrule
\multicolumn{5}{c}{PSNR$\uparrow$} \\
\midrule
SRNs (Single-scene)  & 26.96  & 26.85 & 20.60 & 26.81 \\
NeRF (Single-scene)  & 33.00   & 32.91 & \textbf{28.65} & 36.18 \\
NSVF (Single-scene)  & \bf{33.19}  & \textbf{34.27} & 27.93 & \textbf{37.14} \\ \midrule
\textbf{\nickname{} (Multi-scene)}  & 32.49 & 32.02 & 27.76 & 34.92 \\
\midrule
\multicolumn{5}{c}{SSIM$\uparrow$}
\\
\midrule
SRNs (Single-scene) & 0.910  & 0.947 & 0.757 & 0.923  \\
NeRF (Single-scene) & 0.967  &0.980 &0.856 &0.974 \\
NSVF (Single-scene) & 0.968  & {\bf 0.987} & 0.854 & \textbf{0.980} \\ \midrule
\textbf{\nickname{} (Multi-scene)} & \textbf{0.971}  & 0.982 & \textbf{0.866} & 0.975 \\
\midrule
\multicolumn{5}{c}{LPIPS$\downarrow$}
\\
\midrule
SRNs (Single-scene) & 0.106   & 0.063 & 0.299  & 0.100 \\
NeRF (Single-scene) & 0.046   & 0.028 & 0.206 & 0.121  \\
NSVF (Single-scene) & 0.043   & \textbf{0.010} & \textbf{0.162} & \textbf{0.025}  \\ \midrule
\textbf{\nickname{} (Multi-scene)}  & \textbf{0.032} & 0.019 & 0.167 & 0.040 \\
\bottomrule
\end{tabular*}
\end{table*}

\begin{table*}[ht]
\tabcolsep=0.cm
\small
\centering
\caption{The PSNR, SSIM and LPIPS scores of our \nickname{} and NeRF on four novel scenes of Synthetic-NeRF in Group 1\&2 experiments in Section \ref{sec:exp_genScene}.
\label{table:synNeRF_4scenes_generalization2}}
\begin{tabular*}{1.\textwidth}{@{\extracolsep{\fill}}lccccc}
\toprule
 &  Drums  & Lego & Materials & Ficus & \textit{mean} \\ \midrule
\multicolumn{6}{c}{PSNR$\uparrow$} \\ \midrule
\textbf{\nickname{} (Group 1)}  & 13.23  & 13.53 & 12.26 & 15.47 & 13.62 \\ \midrule
NeRF (Group 2, 100 iters)  & 14.54  & 14.92 & 15.42 & 15.72 & 15.15 \\
NeRF (Group 2, 1k iters)  & 18.01   & 20.04 & 20.40 & 20.81 & 19.81 \\
NeRF (Group 2, 10k iters)  & 21.57  & 24.99 & 23.36 & 23.47 & 23.35 \\ \midrule
\textbf{\nickname{} (Group 2, 100 iters)}  & 18.70 & 20.24 & 18.81 & 21.03 & 19.69 \\
\textbf{\nickname{} (Group 2, 1k iters)}  & 20.49 & 23.64 & 21.87 & 22.02 & 22.00 \\
\textbf{\nickname{} (Group 2, 10k iters)}  & 23.11 & 27.07 & 25.11 & 25.11 & 25.10 \\ \midrule
\multicolumn{6}{c}{SSIM$\uparrow$} \\ \midrule
\textbf{\nickname{} (Group 1)}  & 0.762  & 0.736 & 0.703 & 0.849 & 0.763 \\ \midrule
NeRF (Group 2, 100 iters)  & 0.769  & 0.717 & 0.716 & 0.808 & 0.752 \\
NeRF (Group 2, 1k iters)  & 0.793   & 0.775 & 0.812 & 0.857 & 0.809 \\
NeRF (Group 2, 10k iters)  & 0.865  & 0.862 & 0.877 & 0.896 & 0.875 \\ \midrule
\textbf{\nickname{} (Group 2, 100 iters)}  & 0.822 & 0.813 & 0.829 & 0.878 & 0.835 \\
\textbf{\nickname{} (Group 2, 1k iters)}  & 0.856 & 0.877 & 0.878 & 0.894 & 0.876 \\
\textbf{\nickname{} (Group 2, 10k iters)}  & 0.901 & 0.924 & 0.913 & 0.923 & 0.916 \\ \midrule
\multicolumn{6}{c}{LPIPS$\downarrow$} \\ \midrule
\textbf{\nickname{} (Group 1)}  & 0.256 & 0.273 & 0.301 & 0.150 & 0.246 \\ \midrule
NeRF (Group 2, 100 iters)  & 0.332  & 0.395 & 0.314 & 0.393 & 0.359 \\
NeRF (Group 2, 1k iters)  & 0.254   & 0.264 & 0.229 & 0.164 & 0.228 \\
NeRF (Group 2, 10k iters)  & 0.157  & 0.154 & 0.128 & 0.110 & 0.137 \\ \midrule
\textbf{\nickname{} (Group 2, 100 iters)}  & 0.196 & 0.203 & 0.159 & 0.117 & 0.169 \\
\textbf{\nickname{} (Group 2, 1k iters)}  & 0.154 & 0.138 & 0.123 & 0.097 & 0.128 \\
\textbf{\nickname{} (Group 2, 10k iters)}  & 0.104 & 0.090 & 0.090 & 0.071 & 0.089 \\ 
\bottomrule
\end{tabular*}
\end{table*}

\subsection{Details of Experimental Results on the real-world dataset (3DScan) \cite{Choi2016} in Section \ref{sec:exp_genScene}}
We select four 360-degree-scanned chair scenes from the challenging real-world 3DScan dataset \cite{Choi2016}. A single model is trained for 100000 iterations on 100 images each of three scenes with the following indices: 00032, 00027, 00279. Then, the model is finetuned with a small number of iterations on the scene 00169 from a sparse set of 50 views. The results below show the generality of the features learned by \nickname{}, and that the model quickly converges to plausible representations of complicated real-world object-based scenes. We can see that it is extremely challenging to obtain high-quality results for complex real-world scenes. We leave it for future work to further improve the generalization capability of GRF.

\setlength\tabcolsep{1.5pt}
\begin{table}[h]
	\centering
	\tabcolsep=0.16cm
      \begin{tabular}{lcccc} \toprule
         & PSNR$\uparrow$ & SSIM$\uparrow$ \\ \midrule
         \textbf{\nickname{} (1k iters)} & 18.80 & 0.640\\
         \textbf{\nickname{} (10k iters)} & 20.19 & 0.662 \\ \bottomrule
      \end{tabular}
	\label{table:real_chair_generalization}
\end{table}

\begin{figure*}[ht]
\setlength{\belowcaptionskip}{ -2 pt}
\centering
   \includegraphics[width=1\linewidth]{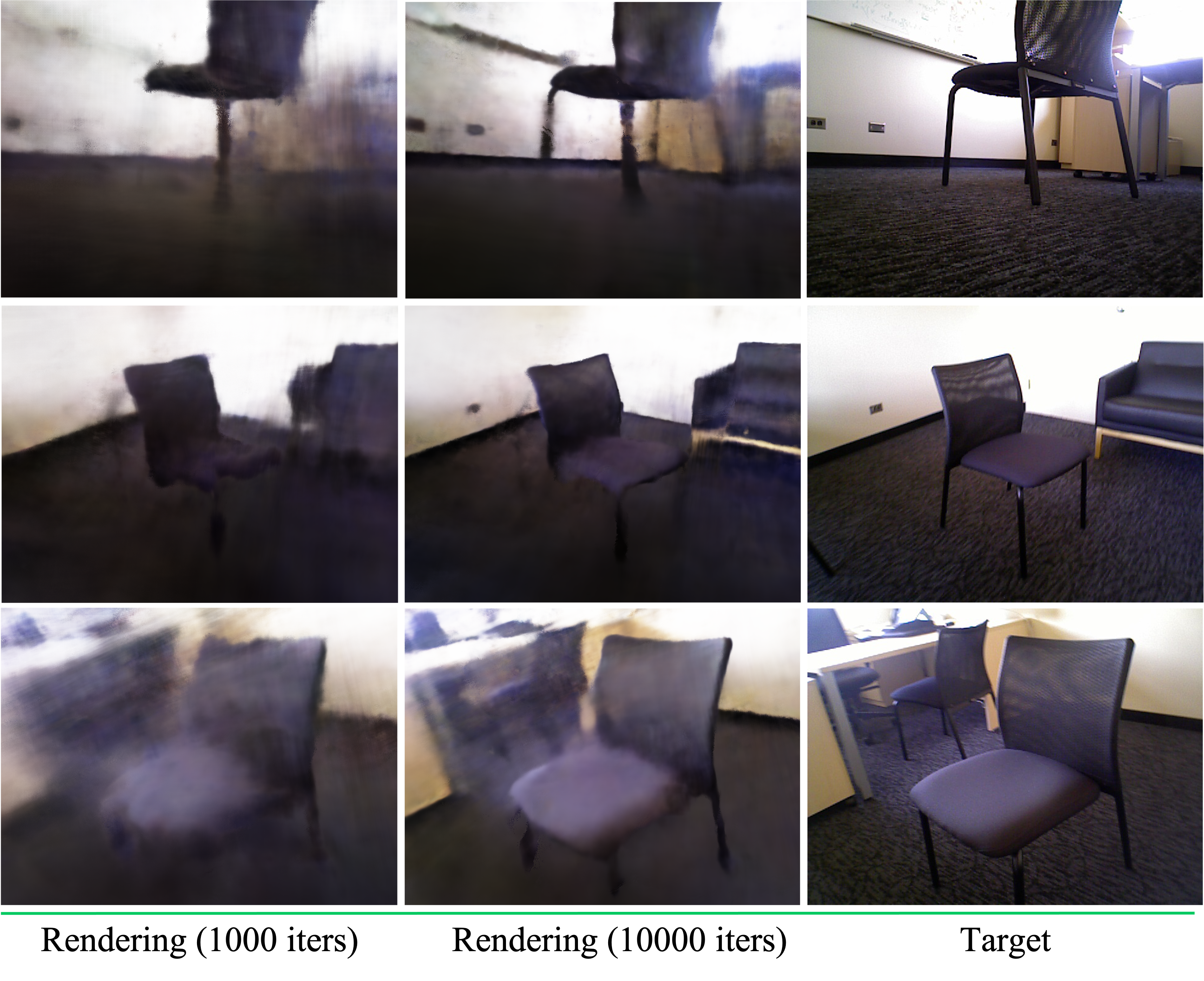}
\caption{Quantitative and Qualitative results of our \nickname{} for novel view synthesis on a real-world chair after finetuning.}
\label{fig:app_realchairs1}
\end{figure*}

\clearpage
\subsection{Details of experimental results on the real-world dataset (LLFF) in Section \ref{sec:exp_singleScene}.}
\begin{table*}[ht!]
\small
\centering
\caption{Comparison of the PSNR, SSIM and LPIPS scores of our \nickname{}, SRNs \cite{Sitzmann2019}, LLFF \cite{Mildenhall2019} and NeRF \cite{Mildenhall2020} in the real-world dataset for single-scene learning in Section \ref{sec:exp_singleScene}.
\label{table:llff_detail}}
\begin{tabular*}{1.\textwidth}{@{\extracolsep{\fill}}lccccccccc}
\toprule
 &  Room &  Fern & Leaves  & Fortress & Orchids & Flower & T-Rex & Horns & \textit{Mean} \\
\midrule
\multicolumn{9}{c}{PSNR$\uparrow$}
\\
\midrule
SRNs  & 27.29 & 21.37 & 18.24 & 26.63 & 17.37 & 24.63 & 22.87 & 24.33 & 22.84 \\
LLFF  & 28.42 & 22.85 & 19.52 & 29.40 & 18.52 & 25.46 & 24.15 & 24.70 & 24.13\\
NeRF & \textbf{32.70} & 25.17 & 20.92 & 31.16 & 20.36 &27.40 &26.80 & 27.45 & 26.50\\
\textbf{\nickname{}(Ours)}  & 31.74 & \textbf{25.72} & \textbf{21.16} & \textbf{31.28} & \textbf{20.88} & \textbf{27.83} & \textbf{27.01} & \textbf{27.50} & \textbf{26.64}  \\
\midrule
\multicolumn{9}{c}{SSIM$\uparrow$}
\\
\midrule
SRNs  & 0.883 & 0.611 & 0.520 & 0.641 & 0.449 & 0.738 & 0.761 & 0.742 & 0.668  \\
LLFF  & 0.932 & 0.753 & 0.697 & 0.872 & 0.588 & 0.844 & 0.857 & 0.840 & 0.798 \\
NeRF  & 0.948 & 0.792 &0.690 & 0.881 &0.641 &0.827 &0.880 & 0.828 & 0.811\\
\textbf{\nickname{}(Ours)}  & \textbf{0.951} & \textbf{0.827} & \textbf{0.727} & \textbf{0.898} & \textbf{0.667} & \textbf{0.852} & \textbf{0.901} & \textbf{0.873} & \textbf{0.837}  \\
\midrule
\multicolumn{9}{c}{LPIPS$\downarrow$}
\\
\midrule
SRNs &0.240 &0.459 &0.440 &0.453 &0.467 &0.288 &0.298 & 0.376 & 0.378\\
LLFF  & 0.155 &0.247 & \textbf{0.216} &0.173 &0.313 &\textbf{0.174} &0.222 &0.193 & 0.212\\
NeRF & 0.178 & 0.280 & 0.316 & 0.171 & 0.321 & 0.219 & 0.249 & 0.268 & 0.250\\
\textbf{\nickname{}(Ours)}  & \textbf{0.104} & \textbf{0.191} & 0.238 & \textbf{0.127} & \textbf{0.275} & 0.176 & \textbf{0.146} & \textbf{0.169} & \textbf{0.178}  \\
\bottomrule
\end{tabular*}
\end{table*}

\begin{table*}[ht]
\small
\centering
\caption{Comparison of the PSNR (in dB), SSIM and LPIPS \cite{Zhang2018f} scores of our \nickname{}, SRNs \cite{Sitzmann2019}, NV \cite{Lombardi2019}, NeRF \cite{Mildenhall2020} and NSVF \cite{Liu2020a} in the Synthetic-NeRF dataset for single-scene learning.
\label{table:synNeRF_detail}}
\begin{tabular*}{1.\textwidth}{@{\extracolsep{\fill}}lccccccccc}
\toprule
 &  Chair &  Drums & Lego  & Mic & Materials & Ship & Hotdog & Ficus & \textit{Mean} \\
\midrule
\multicolumn{9}{c}{PSNR$\uparrow$}
\\
\midrule
SRNs  & 26.96 & 17.18 & 20.85 & 26.85 & 18.09 & 20.60 & 26.81 & 20.73 & 22.26 \\
NV  & 28.33 & 22.58 & 26.08 & 27.78 & 24.22 & 23.93 & 30.71 & 24.79 & 26.05 \\
NeRF & 33.00 & 25.01 & 32.54 & 32.91 & 29.62 & 28.65 & 36.18 & 30.13 & 31.01 \\
NSVF & 33.19 & 25.18 & 32.29 & \textbf{34.27} & \textbf{32.68} & 27.93 & 37.14 & \textbf{31.23} & 31.74 \\
\textbf{\nickname{}(Ours)}  & \textbf{34.51} & \textbf{25.83} & \textbf{32.92} & 33.94 & 30.91 & \textbf{30.12} & \textbf{37.47} & 30.75 & \textbf{32.06}  \\
\midrule
\multicolumn{9}{c}{SSIM$\uparrow$}
\\
\midrule
SRNs  & 0.910 & 0.766 & 0.809 & 0.947 & 0.808 & 0.757 & 0.923 & 0.849 & 0.846  \\
NV  & 0.916 & 0.873 & 0.880 & 0.946 & 0.888 & 0.784 & 0.944 & 0.910 & 0.893  \\
NeRF  & 0.967 &0.925 &0.961 &0.980 &0.949 &0.856 &0.974 &0.964 & 0.947 \\
NSVF  & 0.968 & 0.931 & 0.960 & {\bf 0.987} & {\bf 0.973} & 0.854 & 0.980 & {\bf 0.973} & 0.953 \\
\textbf{\nickname{}(Ours)} & \textbf{0.981} & \textbf{0.937} & \textbf{0.967} & \textbf{0.987} & 0.963 & \textbf{0.891} & \textbf{0.983} & 0.969 & \textbf{0.960}  \\
\midrule
\multicolumn{9}{c}{LPIPS$\downarrow$}
\\
\midrule
SRNs & 0.106 & 0.267 & 0.200 & 0.063 & 0.174 & 0.299 & 0.100 & 0.149  & 0.170\\
NV  & 0.109 & 0.214 & 0.175 & 0.107 & 0.130 & 0.276 & 0.109 & 0.162  & 0.160\\
NeRF & 0.046 & 0.091 & 0.050 & 0.028 & 0.063 & 0.206 & 0.121 & 0.044  & 0.081 \\
NSVF  & 0.043 & 0.069 & \textbf{0.029} & \textbf{0.010} & \textbf{0.021} & 0.162 & \textbf{0.025} & \textbf{0.017} & \textbf{0.047}  \\
\textbf{\nickname{}(Ours)}  & \textbf{0.021} & \textbf{0.068} & 0.042 & 0.013 & 0.041 & \textbf{0.141} & 0.028 & 0.032 & 0.048  \\
\bottomrule
\end{tabular*}
\end{table*}

In order to push the boundaries of single-scene learning, we also conduct experiments on the Synthetic-NeRF dataset in addition to the experiments on real-world scenes in Section \ref{sec:exp_singleScene}. The detailed results are shown in Table \ref{table:synNeRF_detail}. Our \nickname{} outperforms the state-of-the-art NSVF approach on both PSNR and SSIM.

\clearpage
\subsection{Analysis of Attention Mechanism}
The attention mechanism in our \nickname{} aims to automatically select the correct pixel patch from multiple pixel patches where the light rays intersect at the same query 3D point in space.

In order to investigate how the attention mechanism learns to select the useful information, we retrieve the maximal attention score from the observed multiple pixel patches for analysis. Intuitively, the higher the attention score is assigned to a particular pixel patch, the more important that patch for inferring the novel pixel RGB. In particular, we conduct the following experiment using our GRF model trained on ShapeNetv2 Cars. In this case, the AttSets attention module is used (details are in Table \ref{table:attention_shapenetv2}). Given a query light ray, multiple 3D points are sampled to query the network.
\begin{itemize}[leftmargin=*]
\item We firstly try to find the 3D point which is near the surface according to the predicted volume density for points along the ray through a given pixel, if they exist. Otherwise, we ignore such pixels, making them white.
\item Then we compute the M feature vectors from the input M views for these surface points.
\item Thirdly, the attention masks for those M feature vectors are computed. We identify the view whose sum of the attention mask along the feature axis is greatest as the main contributor for inferring the novel pixel RGB.
\item After querying light rays for each pixel, we obtain a rendered RGB image. At the same time, for each pixel of that image, we select the most important view from the M input views for the surface-intersection point along the ray from the viewpoint through that pixel, according to the maximal attention score. Eventually, we  obtain a Max Attention Map corresponding to the rendered RGB image.
\end{itemize}

\begin{figure*}[h]
\centering
\includegraphics[width=0.85\linewidth]{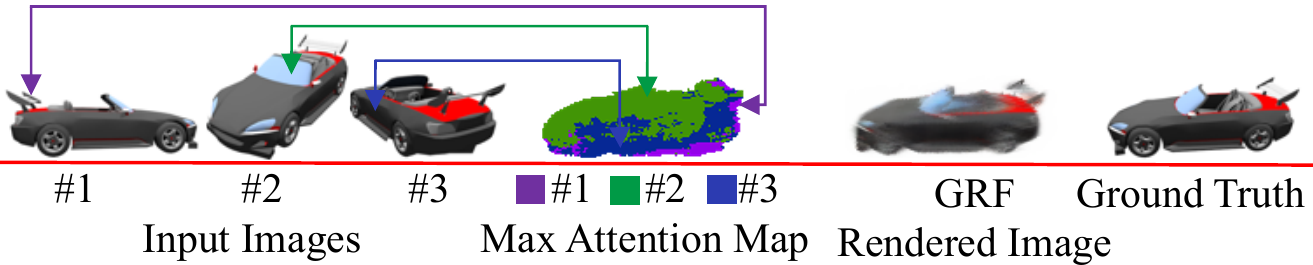}
\caption{Visualization of Max Attention Map from Multiple Input Images of a Novel Object for Inferring a Novel View. }
\label{fig:max_att_map}
\end{figure*}

Figure \ref{fig:max_att_map} shows the qualitative results of the above experiment. In particular, we feed the three images (\#1,\#2,\#3) of an unseen car into our GRF model which is well-trained on car category, and then render a new image (e.g., the 5th image in Figure \ref{fig:max_att_map}). Note that, we carefully select the input 3 images and the rendered image with very large viewing baselines. In the mean time, we obtain and visualize the Max Attention Map corresponding to the rendered image.

For each pixel of the rendered image, we retrieve the input image pixel that has the highest attention score. Specifically, the rendered pixels with purple color correspond to the input image \#1, the green pixels correspond to the input image \#2, while the blue pixels correspond to the input image \#3.

\textbf{Analysis.}
It can be seen that, when inferring a new image, the attention module of our GRF focuses on the most informative pixel patch from the multiple input pixel patches. In addition, it is able to truly deal with the visual occlusion. For example, when inferring the windshield of the car, the attention module focuses on the input image \#2 where the windshield is visible, while ignoring the image \#1 and \#3 where the windshield is self-occluded.

\clearpage
\subsection{Generalization to Visual Occlusions and Variable Input Images}
\label{sec:app_visualOcc}

We carefully select the attention module, i.e., either AttSets or SlotAtt, to aggregate the features from an arbitrary number of input views. In order to evaluate how our \nickname{} is able to generalize with a variable number of input views, especially when there is a very sparse number of views with severe visual occlusions, we conduct the following four groups of experiments.

\begin{itemize}[leftmargin=*]
\item \textbf{1-view Reconstruction}. We feed the a single image of a novel car into our GRF model which is well-trained on car category (trained with 5 images per object), and then render 9 new images from vastly different viewing angles. This is the extreme case where the majority of the object is self-occluded.
\item \textbf{2-view Reconstruction}. Similarly, we feed only two images of the novel car into the same model and render the same 9 novel views. In this case, more information is given to the network, but there are still many parts occluded.
\item \textbf{5-view / 10-view Reconstruction}. The same GRF model is fed with 5 and 10 views of the novel object, rendering the same set of new images.
\end{itemize}

\begin{figure*}[h]
\centering
\includegraphics[width=0.8\linewidth]{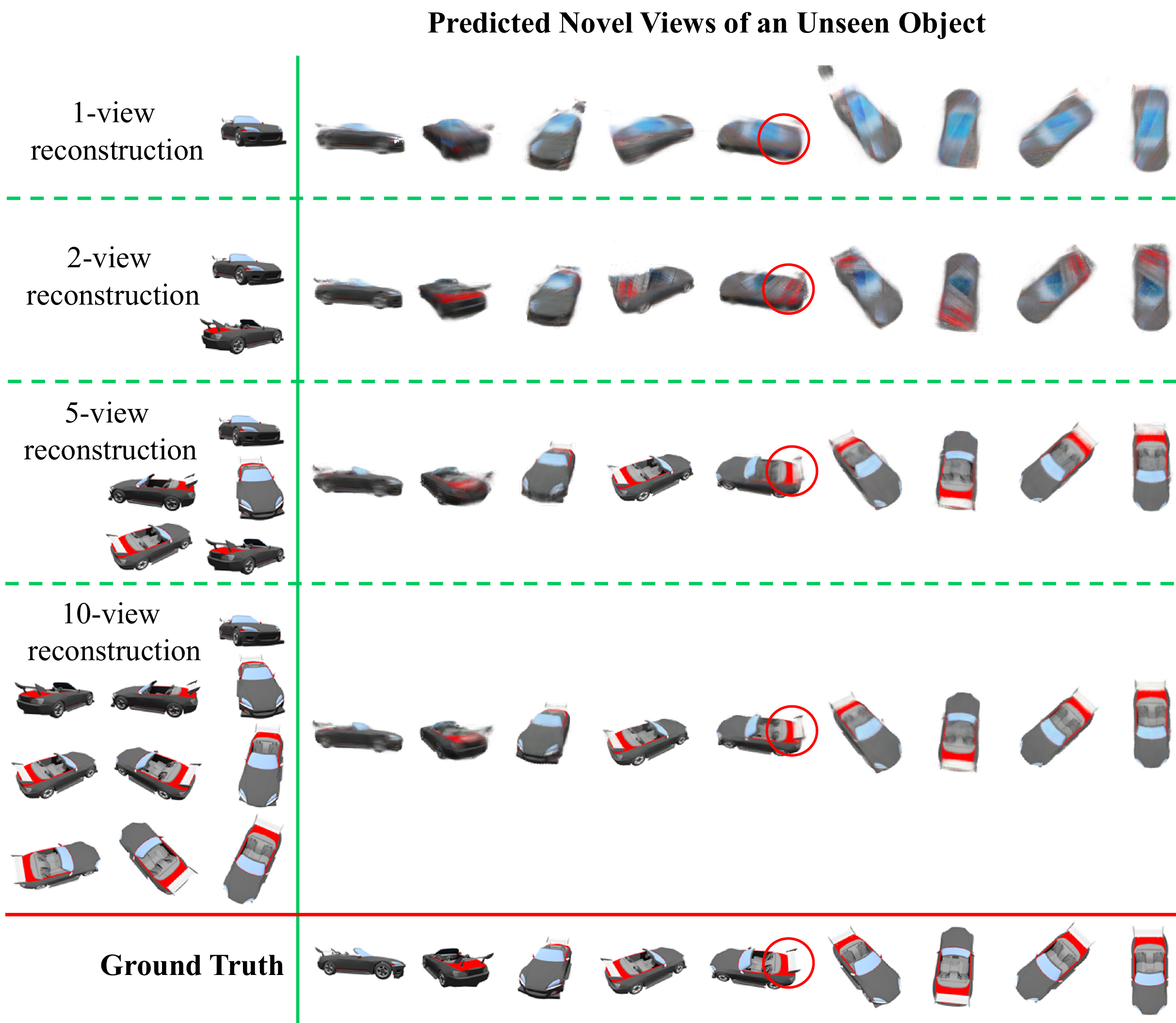}
\caption{Qualitative results of our \nickname{} when being fed with a variable number of views of a novel object. The red circle highlights that the tail of the car is able to be recovered given more visual cues from more input images.}
\label{fig:grf_1_2_5_10_generalization}
\end{figure*}

\textbf{Analysis.} Figure \ref{fig:grf_1_2_5_10_generalization} shows the qualitative results. It can be seen that: 1) In the extreme case, i.e., 1-view reconstruction, our \nickname{} is still able to recover the general 3D shape of the unseen object, including the visually occluded parts, primarily because our CNN model learns the hierarchical features including the high-level shapes. 2) Given more input views, the originally occluded parts tend to be observed from some viewing angles, and then these parts can be reconstructed better and better. This shows that our \nickname{} is indeed able to effectively identify the corresponding useful pixel features for more accurately recovering shape and appearance.

\clearpage
\subsection{More Qualitative Results of real-world scenes in Section \ref{sec:exp_singleScene}}

\begin{figure*}[ht]
\setlength{\belowcaptionskip}{ -2 pt}
\centering
   \includegraphics[width=1\linewidth]{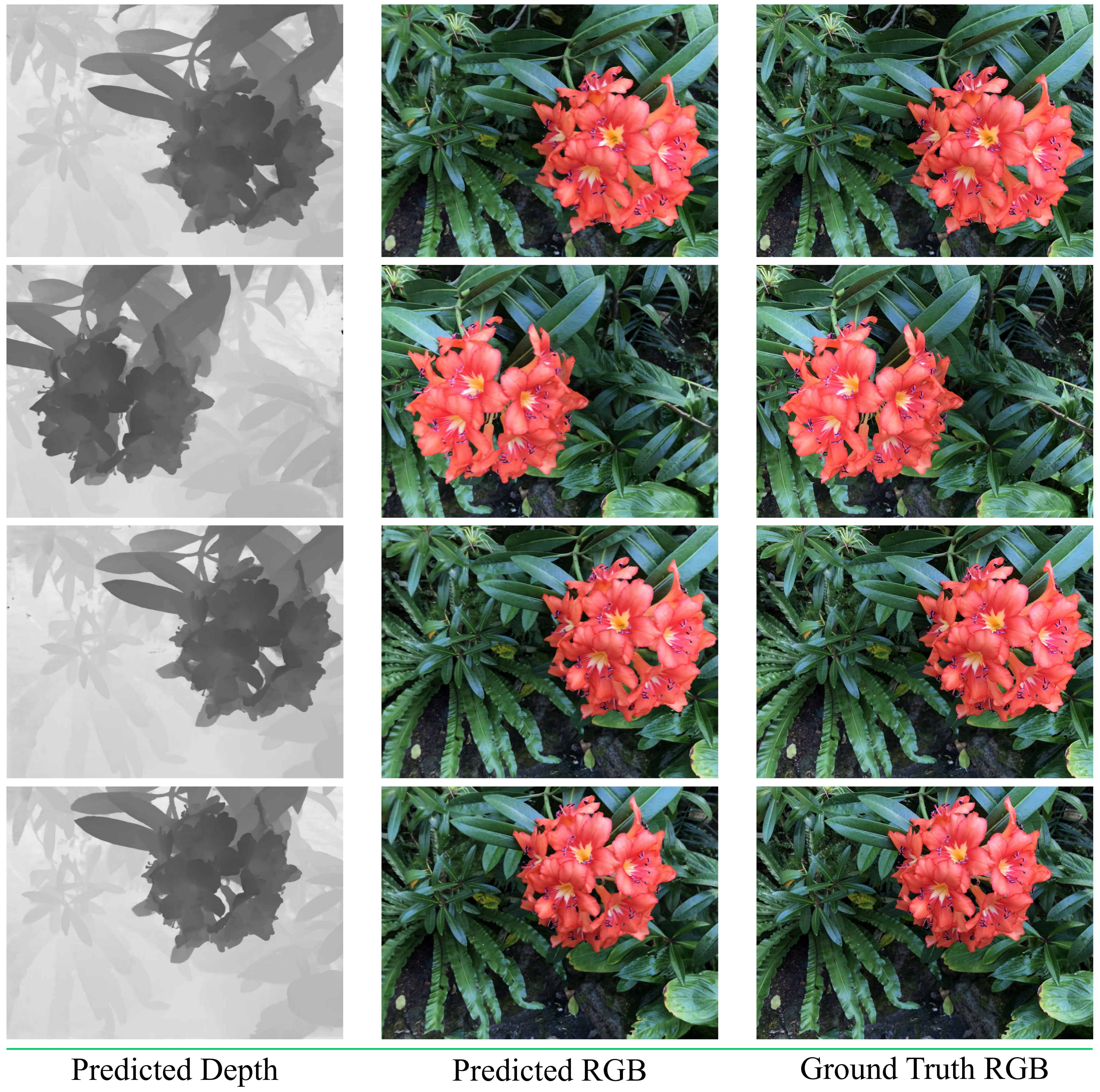}
\caption{Qualitative results of our \nickname{} for novel view depth and RGB estimation on the real-world dataset in Section \ref{sec:exp_singleScene}.}
\label{fig:app_realworld1}
\end{figure*}

\begin{figure*}[ht]
\setlength{\belowcaptionskip}{ -2 pt}
\centering
   \includegraphics[width=1\linewidth]{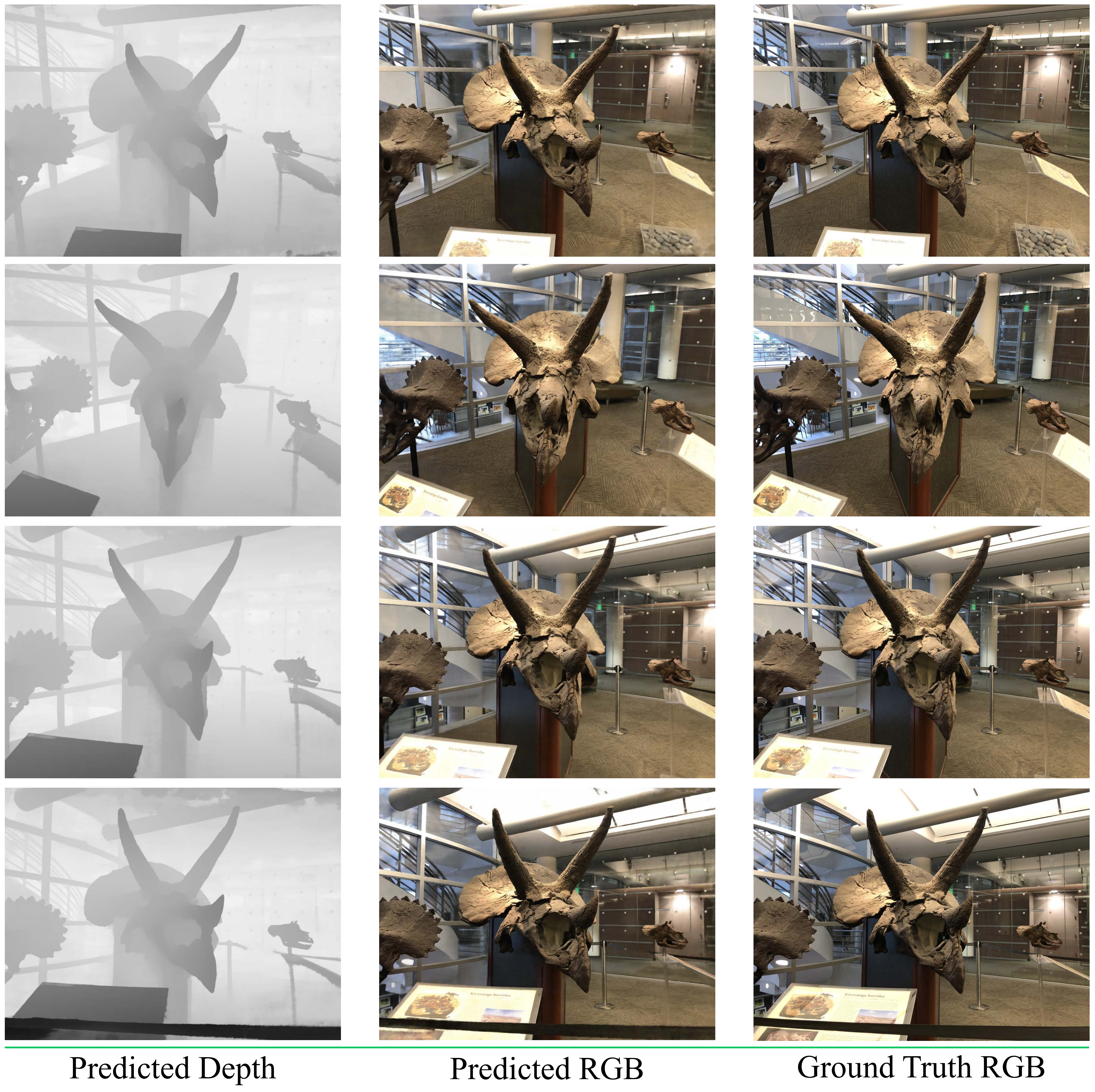}
\caption{Qualitative results of our \nickname{} for novel view depth and RGB estimation on the real-world dataset in Section \ref{sec:exp_singleScene}.}
\label{fig:app_realworld2}
\end{figure*}

\end{document}